# A 6-DOF haptic manipulation system to verify assembly procedures on CAD models

Paolo Tripicchio, Carlo Alberto Avizzano, Massimo Bergamasco

*TeCIP Institute, Scuola Superiore Sant'Anna, Italy*

**Abstract**

During the design phase of products and before going into production, it is necessary to verify the presence of mechanical plays, tolerances, and encumbrances on production mockups. This work introduces a multi-modal system that allows verifying assembly procedures of products in Virtual Reality starting directly from CAD models. Thus leveraging the costs and speeding up the assessment phase in product design. For this purpose, the design of a novel 6-DOF Haptic device is presented. The achieved performance of the system has been validated in a demonstration scenario employing state-of-the-art volumetric rendering of interaction forces together with a stereoscopic visualization setup.

## 1. Introduction

Nowadays the research on virtual environments is carried out in several different fields but only a few among them had a recognized commercial success. In particular the cost, still expensive, of such systems limits their use in fields where the ratio between cost and benefit is relevant. Between them, we can find in short: medicine, military, and industrial applications.

In this latter field, thanks to the high production volumes, the cost of the adoption of a virtual reality system in different phases of the production stage becomes particularly interesting. The advantages introduced by virtual reality to the industrial environment not only regard the formation and staff training but also allow to carry out, before the production phase, a set of functional tests as well as to perform evaluation and correction of virtual prototypes, to analyze the ergonomics of the products and even to realize procedures of maintenance and remote assistance[8].

The most of the virtual reality systems, however, suffer from intuitive interaction and manipulation lacks so that an important role is played by the discrepancy among the level of interaction required and the ability offered by the technologies[23].

With the multi-modal system presented in the following, it is possible to achieve an added value in the production process. In particular, the system allows recognizing, during the simulation, assembly problems directly on the design phase without the need for expensive prototypes or mock-ups. In the same time, it is possible to verify the stresses and strains acting on the mechanical components during the normal usage and the assembly phase. A volumetric haptic rendering approach grants high performance of the system with a realistic simulation of the interaction forces during the assembly stage.

Virtual reality systems were adopted in the past for assembly planning [2] with the goal of determining correct sequences to assemble a final product. The training of the whole assembly process has been studied to enhance the learning phase in [6]. Particular algorithms are implemented in [19] to aid the simulation of model assemblies by means of constraint guidance. In [11] path planning algorithms are studied to assist the user during assembly/disassembly tasks. Physical prototypes have been successfully substituted by virtual counterparts with good results [16]. In particular, the introduction of haptic feedback in the prototyping phase allows designers to make better decision [21].

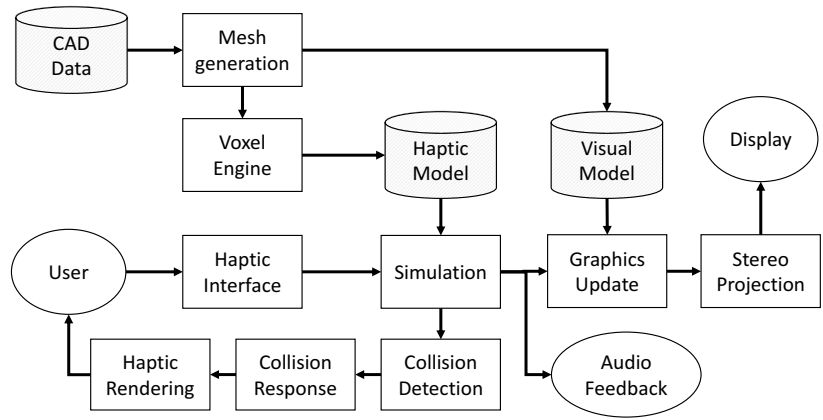

Figure 1: Schematic representation of the modules composing the system architecture.

The objectives of the work were to design a multi-modal interaction system provided with audio, visual and haptic feedback to be employed in the industrial production cycle with the aim of testing assembly procedures requiring a realistic display of the interaction forces and collisions. The solution presented shows a particular serial/parallel manipulator integration that allows to display large workspaces while maintaining a good manipulability indexes, when compared with traditional parallel solutions[22], but also to achieve small encoumbrance with high translational/rotational rigidity, when compared to cable-driven haptics[9] and serial ones (one for all [12]).

Particular attention was given to maintain a coherent multi-sensory integration[4] during the simulation providing collocated perception by means of an immersive and stereographic vision system, a real-time synthesis of audio effects (simulating impact sounds between mechanical parts) and allow full manipulandum capabilities (6DOF) with a force feedback interface. Moreover, the choice of the haptic device was subject to an ergonomic manipulation design criterion and the haptic rendering was intended for mediated interaction with accurate modeling of physical interaction.

Section 2 will introduce the overall system setup, its architecture and the visualization middleware adopted for the graphical rendering. Section 3 will introduce the haptical interfaces of the system and in section 4 the volumetric rendering approaches are detailed. In section 5 an usage scenario is presented. Finally section 6 will comment about the present and future works on the topic.

## 2. Architecture

The system, object of this work, is composed of several modules. A shared communication layer is responsible for the information exchange and the proper functioning of the system itself. The architecture of the system is shown in Figure 1 and its main components can be summarized as:

- a Virtual Reality (VR) Scene with Stereographic projection
- a Haptic Interface (HI) with high force resolution and good manipulation properties
- a Volumetric haptic rendering and collision detection algorithm
- a Distributed computing system with a UDP network communication layer

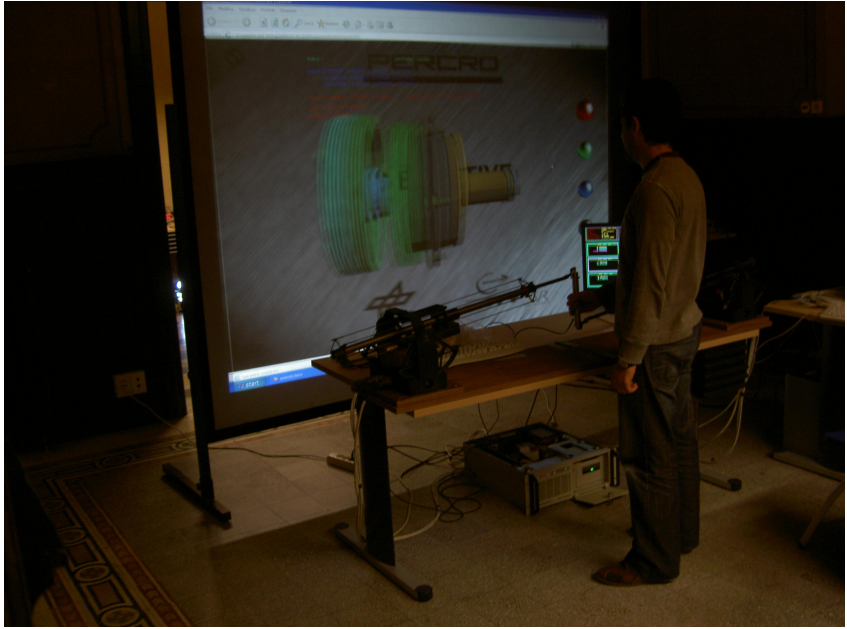

Figure 2: A user during the execution of a virtual assembly task.

The haptic software system incorporates visual, auditory and haptic feedback. The haptic subsystem is generally composed by a collision detection algorithm and a haptic rendering part that feeds back the calculated force. A synchronization engine then links the haptic subsystem with feedback provided by the Audio/Video subsystem, this last is responsible for loading (importing) virtual reality objects from the virtual world and render them on a stereographic display. For each virtual object, a corresponding physical representation is provided for contact and simulation purposes.

The internal system communication have been implemented through two different modalities:

- Support for direct APIs (+libraries): which is being limited to small portion of high performance code especially for the components that communicate with physical hardware. To prevent program complexity, most of the system however implements a distributed (communication) protocol.

- Communication protocols: splits the computing system in several standalone programs each one implementing a specific functionality. The modules talk to each other using standardized communication protocols as well as common remote procedure calls. A set of simple simple data structures were implemented for all relevant inter-process communication (for visualization, sound, haptic device driver, haptic rendering, ...).

All programs communicate via UDP network packets if belonging to different machines or shared memory if the programs run on the same computer. The use of the UDP protocol allows using different OSes on the computers being part of the network without restricting the usable platforms.

The whole system, employed during the testing sessions, spans over three different computing nodes and use a dedicated high-speed ethernet. One host is responsible for the control of the haptic device. A second host runs the haptic rendering algorithms and the third machine serves as the Main user interface, responsible for the visualization, audio and global managing of the system.

Due to performance issues and direct hardware control, instead of using/integrating ROS[17], Unity [3], YARP[15] or other middlewares for robotics, we selected a simpler engine for 3D audio/video rendering, the eXtreme Virtual Reality(XVR) framework[5], which integrates easy management of virtual reality applications with user extension of any type, such those to support custom made hardware and distributed/parallel computing. In our case, we implemented an UDP protocol to allow data sharing across different OSes and real-time performances (>1KHz) are necessary for the display of haptic cues to the users. Local communications instead employ secure shared memory across different modules.

XVR framework benefits include, in addition to those already mentioned abovem the capability to seamless switch rendering from tablet developement to a large set of fully immersive scenarios such as head mounted displays (HMDs, e.g. Oculus Rift), Powerwalls or CAVE[7] displays.

The setup adopted in this work was implemented using projection screen with passive stereoscopic technology by Infitec (See figure 2). This kind of technology makes use of special filters that separate the images for the user's left and right eye respectively applying a separation on the perceived colors (wavelength multiplex visualization). This technique is called the interference filter and is proprietary of the Infitec brand. In the situation in which more than one user wants to use the system, a multi-stereographic approach can be adopted according to the study in [20].

## 3. The Novel 6DOF Device

To achieve the required manipulation capabilities that allows to handle objects in a large workspace while still having excellent haptic response all over the wotkspace, we decided to modify and integrate a couple of existing 3-DOF haptic manipulators that were designed for haptic exploration in large workspaces (the GRAB Manipulators, presented in [1]). Figure 3 shows a schematic of each arm mechanics.

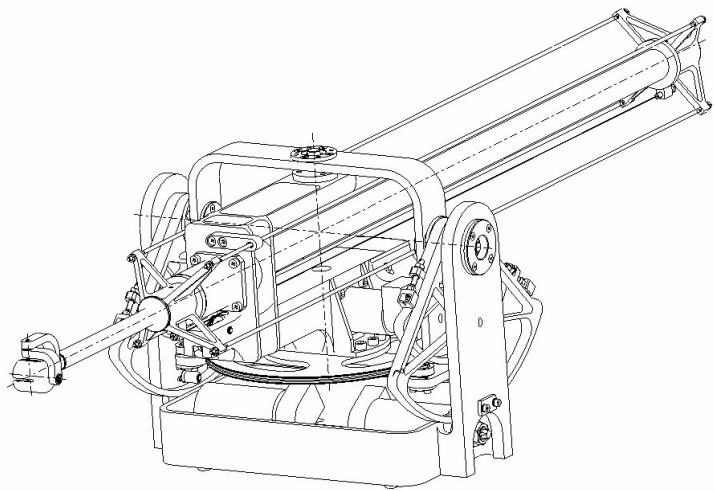

Figure 3: The structure of the GRAB interface

Each device has a total of three actuated Degrees of Freedoms (DOF) in the parallel serial configuration 2R+1P, plus three passive (3R) rotational degrees of mobility (DOM). The two orthogonal rotational pairs works in parallel to control the orientation of a plate to which the third prismatic joint is attached. A balance weight is also placed on the rotating plate on each device in order to minimize the motor efforts when the prismatic joint is fully extended. The first three joints provide to move and track the end-effector (EE) in a 3D space and provide force feedback up 5N in the worst case condition, the remaining three DOM allow to orient the EE in any direction.

The workspace shape is a square based pyramid trunk bigger when the prismatic joint is extended, from which we only use a Rectangular parallelepiped of size 400x400x600mm. GRAB is provided with encoders by which the motor controller can resolve spatial position up to 100um in the worst case condition. The controller also handles seamlessly basic functionality such as gravity compensation, initial calibration, acoustic warnings, and sharp motion protection as in the case of programming/communication errors. GRAB device has very good transparency properties due to different design choices like: remote (base) localization of all motors; use of tendon transmission; absence of gearboxes and use of capstans for principal motions that zero the backlash at joints.

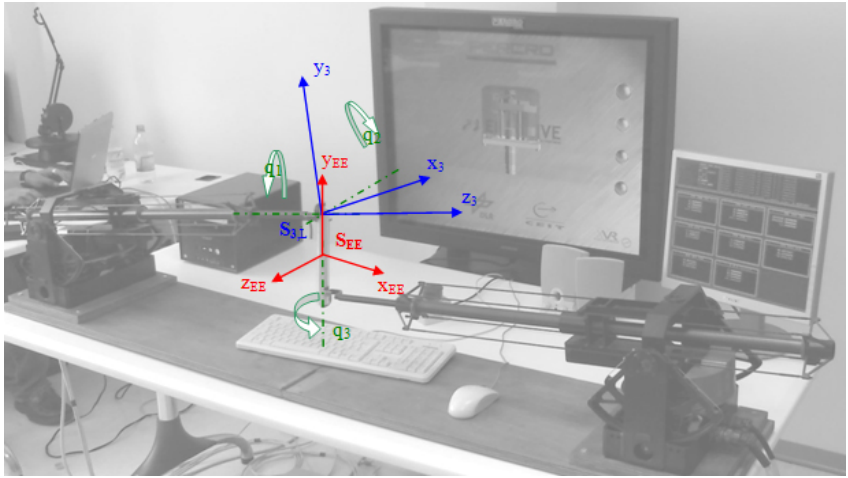

Figure 4: Handle frame and rotation angles of gimbals

In order to extend the manipulability capabilities of this system a custom (motorized and sensorized) handle that interconnects two GRAB devices was developed. The two devices were placed one in front of the other as shown in figure 4. The distance between the devices was set in order to precisely align the operational workspaces of the two devices. This choice was driven by the following motivations:

- Workspace: The design had to guarantee large mobility to match the mobility of the human arm.

- Interference: because of the parallelism of the kinematics the two robotics arms were developed to not interfere in the mechanical structures.

- Force: the force workspace itself should emulate the hand multisensorial receptors that are able to recognize nearly 2 decades in terms of maximum force/force resolution

The setup choice allowed us to achieve a parallel configuration without introducing negative effects on the overall workspace. Moreover the overall system took advantage of two benefits:

- improved force capabilities: the overall worst case condition force more than doubled ($5N \rightarrow 12N$, even considering the not-balance 0.2Kg weight of the coupling device) thanks to the fact that worst case condition on one device now correspond to best force condition on the other (See figure 5);

- improved stiffness: by the combination of the parallel structure with individual device stiffness that improved when the other decreases.

The chosen configuration also improved the isotropy of the device (e.g. the ratio between the min torque and the max torque exerted by the actuators for the same required force at the end-effector and for different working points in the workspace of the device) w.r.t. other kinematics solutions, like that used for the commercial PHANTOM device. This is important in order to obtain a uniform use of the actuators in the device's workspace and to have a uniform reflected inertia.

The coupling element allows the system to display both forces and torques up to 5 DOF. To cope with the torque along the major axis of the coupling handle an additional motor was inserted within the handle. Using the device, the user could take the handle and manipulate objects in the virtual reality scene with great realism. The overall system workspace is similar to the space which can be covered by human arm while sit, thus not introducing relevant limitation to manipulation. Furthermore the control software was

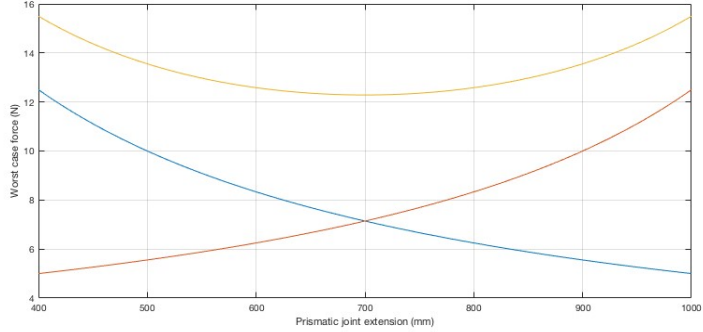

Figure 5: Coupled device force. Red: right device worst case minimum vertical force, Blue: left device worst case minimum vertical force; Yellow: combined device worst case minimum vertical force

modified in order to reach optimal coherence and collocation issues, in particular, different aspects were considered:

We modified the control software to use three different coordinate systems: two local frames relative to the right and the left arms and an independent frame. All the calculated forces and torques are than feedback in the center of mass of the grasping handle and thus are correctly provided to the user for interaction purpose.

3.1. Kinematic Analysis

The kinematics algorithm allows finding the vector position $P_{EE}^0$ and the orientation matrix $R_{EE}^0$ of the handle frame $\Sigma_{EE}$ with respect to the main frame $\Sigma_0$ . The origin of the handle frame is placed on the center of the handle, with the Y-axis coincident with the external cylindrical surface. Input data of the algorithm are the vectors positions ($P_{3,R}^0$ ,$P_{3,L}^0$ ) of the center of gimbals and also the rotation matrices ($R_{3,R}^0$ ,$R_{3,L}^0$ ) of the barrels for the right and left arms; this data are obtained by the direct kinematics of the two arms. The handle frame $\Sigma_{EE}$ , the barrel frame of left arm $\Sigma_{3,L}$ and the angular variables of gimbals $q_i$ are shown in figure 4. The vector position $P_{EE}^0$ , with respect to the main reference frame $\Sigma_0$ , is given by

$$P_{EE}^0 = \frac{P_{3,R}^0 + P_{3,L}^0}{2} \tag{1}$$

In order to obtain the rotation matrix

$$R_{EE}^0 = [X_{EE}^0, Y_{EE}^0, Z_{EE}^0] \tag{2}$$

it is necessary to find the Y axis with respect to the main frame

$$Y_{EE}^0 = \frac{P_{3,R}^0 - P_{3,L}^0}{\|P_{3,R}^0 - P_{3,L}^0\|} \tag{3}$$

and also related to the frame of barrel $\Sigma_3$

$$Y_{EE}^3 = (R_3^0)^T Y_{EE}^0 \tag{4}$$

The direct kinematics of gimbals allows finding the rotation matrix from $\Sigma_{EE}$ to $\Sigma_3$

$$R_{EE}^3 = [X_{EE}^3, Y_{EE}^3, Z_{EE}^3] \tag{5}$$

$$R_{EE}^3 = \begin{bmatrix} c_1 s_2 c_3 - s_1 s_3 & -c_1 c_2 & c_1 s_2 s_3 + s_1 c_3 \\ s_1 s_2 c_3 + c_1 s_3 & -s_1 c_2 & s_1 s_2 s_3 - c_1 c_3 \\ c_2 c_3 & s_2 & c_2 s_3 \end{bmatrix} \tag{6}$$

where $c_x = \cos(q_x)$ and $s_x = \sin(q_x)$.

Combining the two previous expressions, its possible to find the angular rotation for the first two joints of gimbals

$$\begin{cases} q_1 = \tan^{-1}\left(\frac{Y^3_{EE,2}}{Y^3_{EE,1}}\right) \\ q_2 = \sin^{-1}\left(Y^3_{EE,3}\right) \end{cases} \quad (7)$$

while the third is measured by the encoder.
Finally, the rotation matrix $R^0_{EE}$ is given by the following expression

$$R^0_{EE} = R^0_3 R^3_{EE} \quad (8)$$

3.2. Force/Torque Analysis

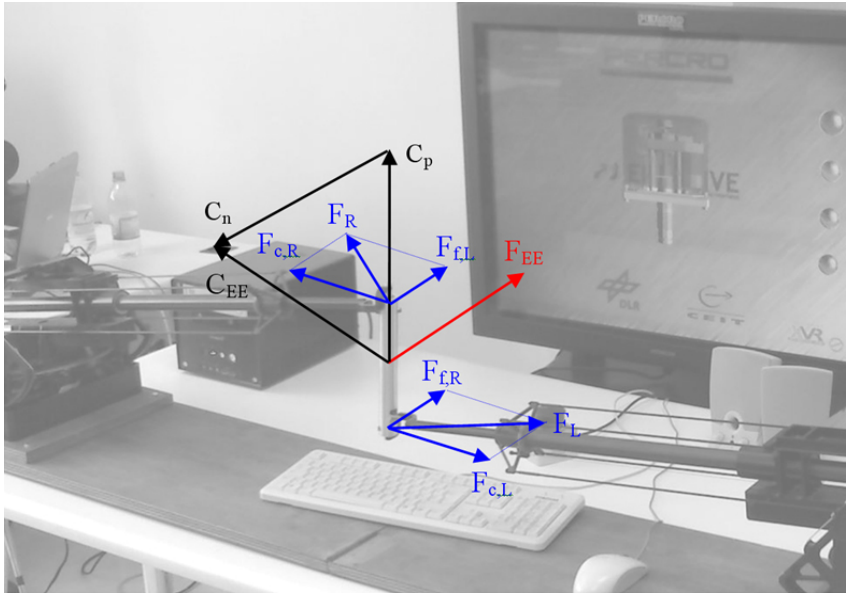

Figure 6: Forces and torques on the handle

The static algorithm allows to find the forces $F^0_R$ and $F^0_L$ that the right and left arms should exert in order to have a given force $F^0_{EE}$ and a torque $C^0_{EE}$ applied on the center of the handle.
The total force that both arms should exert is given by two contributes

$$F^0 = F^0_f + F^0_c \quad (9)$$

The component $F^0_f$ is equal for both arms and it is calculated in order to have the desired force $F^0_{EE}$ on the handle

$$F^0_{f,R} = F^0_{f,L} = \frac{R^3_{EE} F^3_{EE}}{2} \quad (10)$$

The desired torque can be decomposed in two components, one parallel ($C^0_p$) and one normal ($C^0_n$) to the handle

$$C^0_{EE} = C^0_n + C^0_p \quad (11)$$

The two components are given by the following expression

$$\begin{cases} C_p^0 = (C_{EE}^0 \cdot Y_{EE}^0)Y_{EE}^0 \\ C_n^0 = C_{EE}^0 - C_p^0 \end{cases} \quad (12)$$

The parallel component is applied directly by the motor mounted inside the handle. The normal component of the torque is obtained by the two opposite forces applied by the arms:

$$F_{c,R}^0 = -\frac{C_{EE,n}^0 \times Y_{EE}^0}{L} \quad (13)$$

$$F_{c,L}^0 = \frac{C_{EE,n}^0 \times Y_{EE}^0}{L} \quad (14)$$

Table 1 summarizes the characteristics of the 6-DOF Haptic Interface.

| | |
|---|---|
| Workspace | 400x400x600 mm |
| Angular Workspace (Solid angle of movements) | $\approx$ 12 rads |
| Continuous/Peak forces | 12.5N $\rightarrow$ 30N |
| Worst case continuous/peak torque | 1 N/m $\rightarrow$ 2.5N/m |
| Typical continuous/peak torque | 1.8 N/m $\rightarrow$ 10N/m |
| Stiffness Range | 14 - 18 N/mm |
| Worst case Resolution | 100 $\mu m$ |
| Worst case Angular resolution | 0.005 rads |

Table 1: Characteristics of the 6-DOF haptic interface.

## 4. Haptic rendering

The proposed system implements a haptic rendering technique that belongs to the volumetric rendering algorithms originally proposed by McNeely et al. [13]. In this kind of rendering the whole scene is represented by means of a volumetric representation. The base elements being the voxel(cubic volumetric entity), all the 3D elements of the scene have been voxelized, that is from their polygonal mesh representation a new representation comprised of single elementary cubic volumes grouped together has been generated. This kind of representation allows to render really detailed shapes with precision and results in a realistic 6-DOF haptic feedback.

### 4.1. Voxmap-Pointshell algorithm

McNeely et al. applied a discrete collision detection approach to 6-DoF haptic rendering. They separated the virtual tool representation from the rest of the virtual objects' one. The virtual tool object is in fact point sampled while a narrow surface band of the objects belonging to the environment is voxelized. This was necessary to apply a penalty-based force computation. The adopted collision model requires however that deep penetrations do not occur. Later in 2005 [14] the same authors introduced distance fields on the voxel representation with the purpose of estimating the time-to-collision.

The haptic algorithm implemented uses a voxel map as a representation of the static scene and a pointshell for the manipulated object[18]. Thus, collision detection can be performed as simple index operations in the Voxel map. This allows a constant execution rate independent of the complexity of the scene itself. The force calculation is done by integrating the partial forces of each colliding point of the pointshell.

The algorithm works accordingly three basic steps: first, the algorithm determines which points of the pointshell object come into contact with the voxel object; then, collision forces are calculated, based on the penetration depth of the pointshell model into the voxmap along a normal perpendicular to a plane that is tangent to the object surface at each point; finally, collision forces vectors are summed together and numerically integrated. When completed, the generated displacement is fed to a virtual coupling to generate forces at the haptic device.

The haptic models (pointshell, voxel map) has been generated out of 3d polygonal models taken from real production CAD and exported as Inventor or VRML for voxelization purpose. Figure 7 shows the pointshell representations of some mechanical parts. Figure 8 shows the voxelization result of a complex mechanical actuator. In particular, considering classical CAD formats, a mechanical assembly is composed usually by a collection of assembled parts. From these files, for each part in the assembly, a pointshell representation is generated, while to achieve the described interaction, for each subset of parts in the assembly a dedicated voxel model is produced. This process can be run offline, thus not needing the knowledge of a predefined ordering for the sequence of the parts to assemble, or online employing GPU techniques that allow generating voxel assembly in a really short amount of time [10].

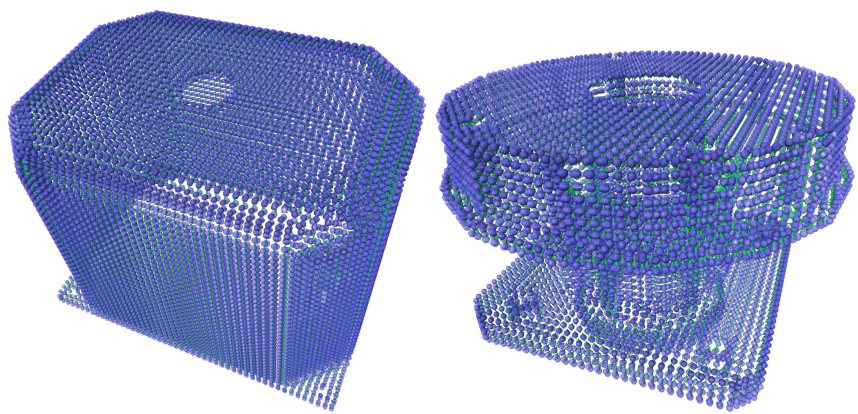

Figure 7: Two mechanical parts represented as pointshells with different spatial resolutions

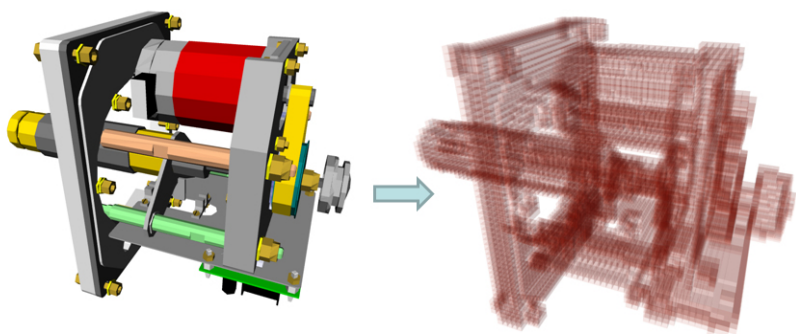

Figure 8: The mechanical actuator mesh on the left is converted to the volumetric representation on the right

## 5. Test scenario

The knowledge of both geometrical properties and constraints of mechanical parts and practical skill are requested in order to accomplish a mechanical assembly task. In this scenario, some mechanical parts taken from a real production CAD assembly are presented to the user for assembling purpose. As the demo begins users can see the final assembled CAD and start to understand how the final piece should look like after the assembling steps. Then, some assembly steps are presented sequentially to the user. In each step, a part is selected as the one to be mounted, and the respective pointshell is used for the interaction. All the previously mounted parts are cooked as a unique voxel map and, for visualization purposes, the desired target configuration of the selected part, rendered in transparency, is displayed onto the virtual model as a visual hint in order to help user interaction. Through the use of senses, the user can perceive and understand the correct sequences of movements to assemble parts together. Visual, audio and haptic cues allow having a good awareness of the interactions of the parts during an assembly operation.

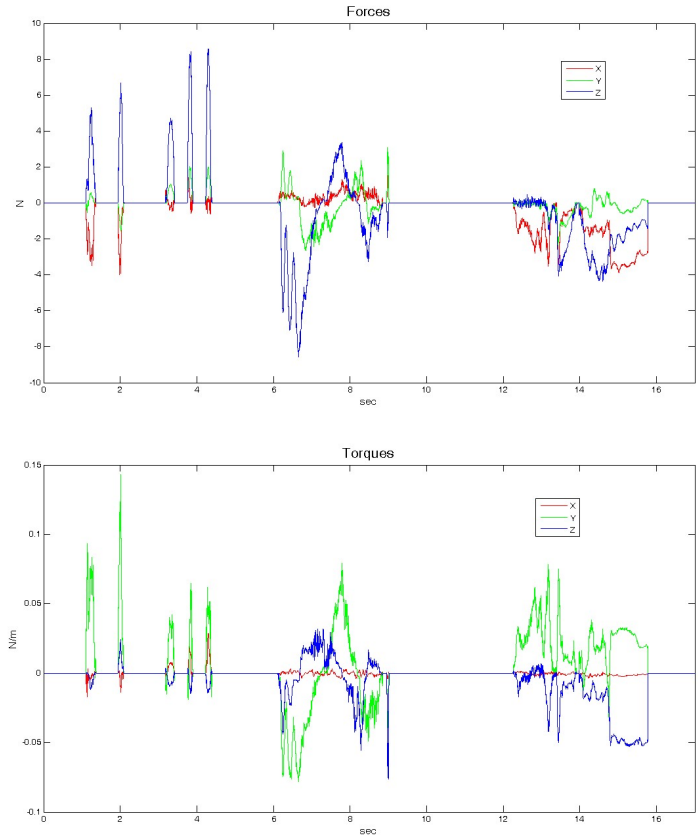

Figure 9: Forces and torques on the haptic interface during an assembly step. Five impacts between the parts and two short interactions are shown.

Overall the test showed a very low resource requirements. During the assembly tests, the haptic algorithm was able to run in real-time at 1600 Hz. The haptic rendering was performed on a low cost architecture, a machine dual core Intel running at 2.33GHz with 2 GB RAM. The VPS algorithm managed an average of 185030 volume entities during the simulated interaction scenario. The network load was also monitored, four kinds of UDP packets were running on the network to exchange positions, orientations, and forces together with configuration parameters. The average network load was measured in 14.5 Mbps. Typical feedback forces and torques exchanged during assembly interaction are shown in Figure 9. The plot have been extracted during one of the assembly steps and shows five rapid collisions (impacts) between the pointshell and the voxmap and two interactions where the pointshell was sliding on the voxmap during a temptative assembly.

## 6. Conclusions and future work

A new multi-modal 6-DOF Haptic device has been developed to achieve good manipulation capabilities. The interface was designed to optimize the realism during mechanical pieces manipulation and to validate assembly procedures directly from production CAD models. A pipeline about how to implement model conversion and rendering was presented, discussed and validated. A test environment provides a multimodal perceptual experience trough Audio, Visual and Haptic feedback. Stereographic Vision immerse the user in the environment granting a realistic three-dimensional interaction. The proposed system allows users to test assembly operations in Virtual reality starting directly from the original CAD designs. It decompose automatically the designs generating virtual representations for haptic interaction. The capabilities of the presented haptic device allows to display in a transparent way good ranges of forces and torques.